\documentstyle[eacl93,named]{article}

\makeatletter
\def\lop#1\to#2{\expandafter\lopoff#1\lopoff#1#2}
\long\def\lopoff,#1,#2\lopoff#3#4{\def#4{#1}\def#3{,#2}}
\def\@@mlistempty{,}
\newif\iflistnonempty
\def\multiputlist(#1,#2)(#3,#4){\@ifnextchar
[{\@imultiputlist(#1,#2)(#3,#4)}{\@imultiputlist(#1,#2)(#3,#4)[]}}

\long\def\@imultiputlist(#1,#2)(#3,#4)[#5]#6{{%
\@xdim=#1\unitlength \@ydim=#2\unitlength
\listnonemptytrue \def\@@mlist{,#6,} 
\loop
\lop\@@mlist\to\@@firstoflist
\@killglue\raise\@ydim\hbox to\z@{\hskip
\@xdim\@imakepicbox(0,0)[#5]{\@@firstoflist}\hss}
\advance\@xdim #3\unitlength\advance\@ydim #4\unitlength
\ifx\@@mlist\@@mlistempty \listnonemptyfalse\fi
\iflistnonempty
\repeat\relax
\ignorespaces}}
\newcount\@@multicnt
\def\matrixput(#1,#2)(#3,#4)#5(#6,#7)#8#9{%
\ifnum#5>#8\@matrixput(#1,#2)(#3,#4){#5}(#6,#7){#8}{#9}%
\else\@matrixput(#1,#2)(#6,#7){#8}(#3,#4){#5}{#9}\fi}

\long\def\@matrixput(#1,#2)(#3,#4)#5(#6,#7)#8#9{{\@killglue%
\@multicnt=#5\relax\@@multicnt=#8\relax%
\@xdim=0pt%
\@ydim=0pt%
\setbox\@tempboxa\hbox{\@whilenum \@multicnt > 0\do {%
\raise\@ydim\hbox to \z@{\hskip\@xdim #9\hss}%
\advance\@multicnt \m@ne%
\advance\@xdim #3\unitlength\advance\@ydim #4\unitlength}}%
\@xdim=#1\unitlength%
\@ydim=#2\unitlength%
\@whilenum \@@multicnt > 0\do {%
\raise\@ydim\hbox to \z@{\hskip\@xdim \copy\@tempboxa\hss}%
\advance\@@multicnt \m@ne%
\advance\@xdim #6\unitlength\advance\@ydim #7\unitlength}%
\ignorespaces}}
\newcount\d@lta
\newdimen\@delta
\newdimen\@@delta
\newcount\@gridcnt
\def\grid(#1,#2)(#3,#4){\@ifnextchar [{\@igrid(#1,#2)(#3,#4)}%
{\@igrid(#1,#2)(#3,#4)[@,@]}}

\long\def\@igrid(#1,#2)(#3,#4)[#5,#6]{%
\makebox(#1,#2){%
\@delta=#1pt\@@delta=#3pt\divide\@delta \@@delta\d@lta=\@delta%
\advance\d@lta \@ne\relax\message{grid=\the\d@lta\space x}%
\multiput(0,0)(#3,0){\d@lta}{\hbox to\z@{\hskip -\@halfwidth \vrule
	 \@width \@wholewidth \@height #2\unitlength \@depth \z@\hss}}%
\ifx#5@\relax\else%
\global\@gridcnt=#5%
\multiput(0,0)(#3,0){\d@lta}{%
\makebox(0,-2)[t]{\number\@gridcnt\global\advance\@gridcnt by #3}}%
\global\@gridcnt=#5%
\multiput(0,#2)(#3,0){\d@lta}{\makebox(0,0)[b]{\number\@gridcnt\vspace{2mm}%
\global\advance\@gridcnt by #3}}%
\fi%
\@delta=#2pt\@@delta=#4pt\divide\@delta \@@delta\d@lta=\@delta%
\advance\d@lta \@ne\relax\message{\the\d@lta . }%
\multiput(0,0)(0,#4){\d@lta}{\vrule \@height \@halfwidth \@depth \@halfwidth
	 \@width #1\unitlength}%
\ifx#6@\relax\else
\global\@gridcnt=#6%
\multiput(0,0)(0,#4){\d@lta}{%
\makebox(0,0)[r]{\number\@gridcnt\ \global\advance\@gridcnt by #4}}%
\global\@gridcnt=#6%
\multiput(#1,0)(0,#4){\d@lta}{%
\makebox(0,0)[l]{\ \number\@gridcnt\global\advance\@gridcnt by #4}}%
\fi}}
\def\picsquare{\hskip -0.5\@wholewidth%
\vrule height \@halfwidth depth \@halfwidth width \@wholewidth}
\def\picsquare@bl{\vrule height \@wholewidth depth \z@  width \@wholewidth}
\newif\if@jointhem \global\@jointhemfalse
\newif\if@firstpoint \global\@firstpointtrue
\newcount\@joinkind
\def\dottedjoin{\global\@jointhemtrue \global\@joinkind=0\relax
  \bgroup\@ifnextchar[{\@idottedjoin}{\@idottedjoin[\picsquare@bl]}}
\def\@idottedjoin[#1]#2{\gdef\dotchar@join{#1}\gdef\dotgap@join{#2}}
\def\enddottedjoin{\global\@jointhemfalse \global\@firstpointtrue\egroup}
\def\dashjoin{\global\@jointhemtrue \global\@joinkind=1\relax
  \bgroup\@ifnextchar[{\@idashjoin}{\@idashjoin[\dashlinestretch]}}
\def\@idashjoin[#1]#2{\edef\dashlinestretch{#1}\gdef\dashlen@join{#2}%
\@ifnextchar[{\@iidashjoin}{\gdef\dotgap@join{}}}
\def\@iidashjoin[#1]{\gdef\dotgap@join{#1}}

\def\drawjoin{\global\@jointhemtrue \global\@joinkind=2\relax
  \bgroup\@ifnextchar[{\@idrawjoin}{}}
\def\@idrawjoin[#1]{\def\drawlinestretch{#1}}

\long\def\jput(#1,#2)#3{{\@killglue\raise#2\unitlength\hbox to \z@{\hskip
#1\unitlength #3\hss}\ignorespaces}
\if@jointhem
 \if@firstpoint \gdef\x@one{#1} \gdef\y@one{#2} \global\@firstpointfalse
 \else\ifcase\@joinkind
	\@dottedline[\dotchar@join]{\dotgap@join\unitlength}%
(\x@one\unitlength,\y@one\unitlength)(#1\unitlength,#2\unitlength)
	\or\@dashline[\dashlinestretch]{\dashlen@join}[\dotgap@join]%
(\x@one,\y@one)(#1,#2)
	\else\@drawline[\drawlinestretch](\x@one,\y@one)(#1,#2)\fi
    \gdef\x@one{#1} \gdef\y@one{#2}
 \fi
\fi}
\newdimen\@dotgap
\newdimen\@ddotgap
\newcount\@x@diff
\newcount\@y@diff
\newdimen\x@diff
\newdimen\y@diff
\newbox\@dotbox
\newcount\num@segments
\newcount\num@segmentsi
\newif\ifsqrt@done
\def\sqrtandstuff#1#2#3{
\ifdim #1 <0pt \@x@diff= -#1 \else\@x@diff=#1\fi
\ifdim #2 <0pt \@y@diff= -#2 \else\@y@diff=#2\fi
\@dotgap=#3 \divide\@dotgap \tw@
\advance\@x@diff \@dotgap \advance\@y@diff \@dotgap
\@dotgap=#3
\divide\@x@diff \@dotgap \divide\@y@diff \@dotgap
\sqrt@donefalse
\ifnum\@x@diff < 2
   \ifnum\@y@diff < 2 \num@segments=\@x@diff \advance\num@segments \@y@diff
		      \sqrt@donetrue
        \else\num@segments=\@y@diff \sqrt@donetrue\fi
   \else\ifnum\@y@diff < 2 \num@segments=\@x@diff \sqrt@donetrue\fi
\fi
\ifsqrt@done \ifnum\num@segments=\z@ \num@segments=\@ne\fi\relax
 \else \ifnum\@y@diff >\@x@diff
		 \@tempcnta=\@x@diff \@x@diff=\@y@diff \@y@diff=\@tempcnta
       \fi    		
  \num@segments=\@y@diff
  \multiply\num@segments \num@segments
  \multiply\num@segments by 457
  \divide\num@segments \@x@diff
  \advance\num@segments by 750 
  \divide\num@segments \@m
  \advance\num@segments \@x@diff
\fi}
\def\dottedline{\@ifnextchar [{\@idottedline}{\@idottedline[\picsquare@bl]}}
\def\@idottedline[#1]#2(#3,#4){\@ifnextchar (%
{\@iidottedline[#1]{#2}(#3,#4)}{\relax}}
\def\@iidottedline[#1]#2(#3,#4)(#5,#6){\@dottedline[#1]{#2\unitlength}%
(#3\unitlength,#4\unitlength)(#5\unitlength,#6\unitlength)%
\@idottedline[#1]{#2}(#5,#6)}
\long\def\@dottedline[#1]#2(#3,#4)(#5,#6){{%
\x@diff=#5\relax\advance\x@diff by -#3\relax
\y@diff=#6\relax\advance\y@diff by -#4\relax
\sqrtandstuff{\x@diff}{\y@diff}{#2}
\divide\x@diff \num@segments
\divide\y@diff \num@segments
\advance\num@segments \@ne     
\setbox\@dotbox\hbox{#1}
\@xdim=#3 \@ydim=#4
\ifdim\ht\@dotbox >\z@
  \advance\@xdim -0.5\wd\@dotbox
  \advance\@ydim -0.5\ht\@dotbox
  \advance\@ydim .5\dp\@dotbox\fi
\@killglue
\loop \ifnum\num@segments > 0
\unskip\raise\@ydim\hbox to\z@{\hskip\@xdim #1\hss}%
\advance\num@segments \m@ne\advance\@xdim\x@diff\advance\@ydim\y@diff%
\repeat
\ignorespaces}}
\def\dashlinestretch{0} 
\def\dashline{\@ifnextchar [{\@idashline}{\@idashline[\dashlinestretch]}}
\def\@idashline[#1]#2{\@ifnextchar [{\@iidashline[#1]{#2}}%
{\@iidashline[#1]{#2}[\@empty]}} 
\def\@iidashline[#1]#2[#3](#4,#5){\@ifnextchar (%
{\@iiidashline[#1]{#2}[#3](#4,#5)}{\relax}}
\def\@iiidashline[#1]#2[#3](#4,#5)(#6,#7){%
\@dashline[#1]{#2}[#3](#4,#5)(#6,#7)%
\@iidashline[#1]{#2}[#3](#6,#7)}
\long\def\@dashline[#1]#2[#3](#4,#5)(#6,#7){{%
\x@diff=#6\unitlength \advance\x@diff by -#4\unitlength
\y@diff=#7\unitlength \advance\y@diff by -#5\unitlength
\@tempdima=#2\unitlength \advance\@tempdima -\@wholewidth
\sqrtandstuff{\x@diff}{\y@diff}{\@tempdima}
\ifnum\num@segments <3 \num@segments=3\fi
\@tempdima=\x@diff \@tempdimb=\y@diff
\divide\@tempdimb by\num@segments
\divide\@tempdima by\num@segments
{\ifx#3\@empty \relax
    \ifdim\@tempdima < 0pt \x@diff=-\@tempdima\else\x@diff=\@tempdima\fi
    \ifdim\@tempdimb < 0pt \y@diff=-\@tempdimb\else\y@diff=\@tempdimb\fi
    \ifdim\x@diff < 0.3pt 
           \ifdim\@tempdimb > 0pt
	        \global\setbox\@dotbox\hbox{\hskip -\@halfwidth \vrule
		 \@width \@wholewidth \@height \@tempdimb}
	   \else\global\setbox\@dotbox\hbox{\hskip -\@halfwidth \vrule
		 \@width \@wholewidth \@height\z@ \@depth -\@tempdimb}\fi
       \else\ifdim\y@diff < 0.3pt 
               \ifdim\@tempdima >0pt
		  \global\setbox\@dotbox\hbox{\vrule \@height \@halfwidth
		 		\@depth \@halfwidth \@width \@tempdima}
		\else\global\setbox\@dotbox\hbox{\hskip \@tempdima
			 \vrule \@height \@halfwidth \@depth \@halfwidth
				 \@width -\@tempdima \hskip \@tempdima}\fi
	    \else\global\setbox\@dotbox\hbox{%
\@dottedline[\picsquare]{0.98\@wholewidth}(0pt,0pt)(\@tempdima,\@tempdimb)}
\fi\fi
\else\global\setbox\@dotbox\hbox{%
\@dottedline[\picsquare]{#3\unitlength}(0pt,0pt)(\@tempdima,\@tempdimb)}
\fi}
\advance\x@diff by -\@tempdima 
\advance\y@diff by -\@tempdimb
\@tempdima=\num@segments\@wholewidth \@tempdima=2\@tempdima 
\@tempcnta=\@tempdima \@tempdima=#2\unitlength \@tempdimb=0.5\@tempdima
\@tempcntb=\@tempdimb \advance\@tempcnta by \@tempcntb 
\divide\@tempcnta by\@tempdima \advance\num@segments by -\@tempcnta
\ifnum #1=0 \relax\else\ifnum #1 < -100
  \typeout{***dashline: reduction > -100 percent implies blankness!***}
\else\num@segmentsi=#1 \advance\num@segmentsi by 100
     \multiply\num@segments by\num@segmentsi \divide\num@segments by 100
\fi\fi
\divide\num@segments by 2 
\ifnum\num@segments >0 
 \divide\x@diff by\num@segments
 \divide\y@diff by\num@segments
 \advance\num@segments by\@ne 
 \else\num@segments=2 
\fi
\@xdim=#4\unitlength \@ydim=#5\unitlength
\@killglue
\loop \ifnum\num@segments > 0
\unskip\raise\@ydim\hbox to\z@{\hskip\@xdim \copy\@dotbox\hss}%
\advance\num@segments \m@ne\advance\@xdim\x@diff\advance\@ydim\y@diff%
\repeat
\ignorespaces}}
\newif\if@flippedargs
\def\lineslope(#1,#2){%
\ifdim #1 <0pt \@xdim= -#1 \else\@xdim=#1\fi
\ifdim #2 <0pt \@ydim= -#2 \else\@ydim=#2\fi
\ifdim\@xdim >\@ydim \@tempdima=\@xdim \@xdim=\@ydim \@ydim=\@tempdima
\@flippedargstrue\else\@flippedargsfalse\fi
\ifdim\@ydim >1pt \@tempcnta=\@ydim
            \divide\@tempcnta by 65536
            \divide\@xdim \@tempcnta\fi
\ifdim\@xdim <.083333pt \@xarg=1 \@yarg=0
 \else\ifdim\@xdim <.183333pt	\@xarg=6 \@yarg=1
 \else\ifdim\@xdim <.225pt 	\@xarg=5 \@yarg=1
 \else\ifdim\@xdim <.291666pt 	\@xarg=4 \@yarg=1
 \else\ifdim\@xdim <.366666pt 	\@xarg=3 \@yarg=1
 \else\ifdim\@xdim <.45pt 	\@xarg=5 \@yarg=2
 \else\ifdim\@xdim <.55pt 	\@xarg=2 \@yarg=1
 \else\ifdim\@xdim <.633333pt 	\@xarg=5 \@yarg=3
 \else\ifdim\@xdim <.708333pt 	\@xarg=3 \@yarg=2
 \else\ifdim\@xdim <.775pt 	\@xarg=4 \@yarg=3
 \else\ifdim\@xdim <.816666pt 	\@xarg=5 \@yarg=4
 \else\ifdim\@xdim <.916666pt 	\@xarg=6 \@yarg=5
       \else			\@xarg=1 \@yarg=1%
\fi\fi\fi\fi\fi\fi\fi\fi\fi\fi\fi\fi
\if@flippedargs\relax\else\@tempcnta=\@xarg \@xarg=\@yarg
			  \@yarg=\@tempcnta\fi
\ifdim #1 <0pt \@xarg= -\@xarg\fi
\ifdim #2 <0pt \@yarg= -\@yarg\fi
}
\newif\if@toosmall
\newif\if@drawit
\newif\if@horvline
\def\drawlinestretch{0} 
\def\drawline{\@ifnextchar [{\@idrawline}{\@idrawline[\drawlinestretch]}}
\def\@idrawline[#1](#2,#3){\@ifnextchar ({\@iidrawline[#1](#2,#3)}{\relax}}
\def\@iidrawline[#1](#2,#3)(#4,#5){\@drawline[#1](#2,#3)(#4,#5)
\@idrawline[#1](#4,#5)}
\def\@drawline[#1](#2,#3)(#4,#5){{%
\x@diff=#4\unitlength \advance\x@diff by -#2\unitlength
\y@diff=#5\unitlength \advance\y@diff by -#3\unitlength
\ifx\@linefnt\tenln \linethickness{0.5pt} \else \linethickness{0.9pt}\fi
\lineslope(\x@diff,\y@diff)
\@toosmalltrue
{\ifdim\x@diff <\z@ \x@diff=-\x@diff\fi
 \ifdim\y@diff <\z@ \y@diff=-\y@diff\fi
 \ifdim\x@diff >10pt \global\@toosmallfalse\fi
 \ifdim\y@diff >10pt \global\@toosmallfalse\fi}
\@drawitfalse\@horvlinefalse
\ifnum#1 <0 \relax\else\@horvlinetrue\fi
\if@toosmall\@horvlinetrue\fi
\if@horvline
 \ifdim\x@diff =0pt \put(#2,#3){\ifdim\y@diff >0pt \@linelen=\y@diff \@upline
 				\else\@linelen=-\y@diff \@downline\fi}%
 \else\ifdim\y@diff =0pt
          \ifdim\x@diff >0pt \put(#2,#3){\vrule \@height \@halfwidth \@depth
				\@halfwidth \@width \x@diff}
		\else \put(#4,#5){\vrule \@height \@halfwidth \@depth
				\@halfwidth \@width -\x@diff}\fi
       \else\@drawittrue\fi\fi 
\else\@drawittrue\fi
\if@drawit
\ifnum\@xarg< 0 \@negargtrue\else\@negargfalse\fi
\ifnum\@xarg =0 \setbox\@linechar%
\hbox{\hskip -\@halfwidth \vrule \@width \@wholewidth \@height 10.2pt
 \@depth \z@}
\else \ifnum\@yarg =0 \setbox\@linechar%
\hbox{\vrule \@height \@halfwidth \@depth \@halfwidth \@width 10.2pt}
\else \if@negarg \@xarg -\@xarg \@yyarg -\@yarg
        \else \@yyarg \@yarg\fi
\ifnum\@yyarg >0 \@tempcnta\@yyarg \else \@tempcnta -\@yyarg\fi
\setbox\@linechar\hbox{\@linefnt\@getlinechar(\@xarg,\@yyarg)}%
\fi\fi
\if@toosmall
  \@dottedline[\picsquare]{.98\@wholewidth}%
(#2\unitlength,#3\unitlength)(#4\unitlength,#5\unitlength)%
\else
\ifnum\@xarg=0\relax\else\ifdim\x@diff >\z@ \advance\x@diff -\wd\@linechar
  \else\advance\x@diff \wd\@linechar\fi\fi
\ifnum\@yarg=0\relax\else\ifdim\y@diff >\z@\advance\y@diff -\ht\@linechar
  \else\advance\y@diff \ht\@linechar\fi\fi
\ifdim\x@diff <\z@ \@x@diff=-\x@diff \else\@x@diff=\x@diff\fi
\ifdim\y@diff <\z@ \@y@diff=-\y@diff \else\@y@diff=\y@diff\fi
\num@segments=0 \num@segmentsi=0
\ifdim\wd\@linechar >1pt
 \num@segmentsi=\@x@diff \divide\num@segmentsi \wd\@linechar\fi
\ifdim\ht\@linechar >1pt
 \num@segments=\@y@diff \divide\num@segments \ht\@linechar\fi
\ifnum\num@segmentsi >\num@segments \num@segments=\num@segmentsi\fi
\advance\num@segments \@ne 
\ifnum #1=0 \relax \else\ifnum #1 < -99
  \typeout{***drawline: reduction <= -100 percent implies blankness!***}
\else\num@segmentsi=#1 \advance\num@segmentsi by 100
     \multiply\num@segments \num@segmentsi
     \divide\num@segments by 100
\fi\fi
\divide\x@diff \num@segments
\divide\y@diff \num@segments
\advance\num@segments \@ne 
\@xdim=#2\unitlength \@ydim=#3\unitlength
\if@negarg \advance\@xdim -\wd\@linechar\fi
\ifnum\@yarg <0 \advance\@ydim -\ht\@linechar\fi
\@killglue
\loop \ifnum\num@segments > 0
\unskip\raise\@ydim\hbox to\z@{\hskip\@xdim \copy\@linechar\hss}%
\advance\num@segments \m@ne\advance\@xdim\x@diff\advance\@ydim\y@diff%
\repeat
\ignorespaces
\fi
\fi}}
\long\def\splittwoargs#1 #2 {(#1,#2)}
\newif\if@stillmore
\newread\@datafile
\long\def\putfile#1#2{\openin\@datafile = #1
\@stillmoretrue
\loop
\ifeof\@datafile\relax\else\read\@datafile to\@dataline\fi
\ifeof\@datafile\@stillmorefalse
\else\ifx\@dataline\@empty \relax
     \else
\expandafter\expandafter\expandafter\put\expandafter\splittwoargs%
\@dataline{#2}
     \fi
\fi
\if@stillmore
\repeat
\closein\@datafile
}

\title{Similarity between Words \\
       Computed by Spreading Activation on an English Dictionary}

\author{
\begin{minipage}{80mm}
  \begin{center}\large
  {\bf  Hideki Kozima} \\
  Course in Computer Science \\
  and Information Mathematics, \\
  Graduate School, \\
  University of Electro-Communications \\
  1--5--1, Chofugaoka, Chofu, \\
  Tokyo 182, Japan \\
  ({\tt xkozima@phaeton.cs.uec.ac.jp})
  \end{center}
\end{minipage}
\begin{minipage}{80mm}
  \begin{center}\large
  {\bf  Teiji Furugori} \\
  Department of Computer Science \\
  and Information Mathematics, \\
  University of Electro-Communications \\
  1--5--1, Chofugaoka, Chofu, \\
  Tokyo 182, Japan \\
  Tel. +81--424--83--2161 \ (ex.4461) \\
  ({\tt furugori@phaeton.cs.uec.ac.jp})
  \end{center}
\end{minipage}}

\begin{document}

\maketitle

\bibliographystyle{named}

%

\newlength{\lma}\setlength{\lma}{ 5mm}
\newlength{\lmb}\setlength{\lmb}{ 8mm}
\newlength{\lmc}\setlength{\lmc}{ 3mm}

\setlength{\unitlength}{1mm}

%

\newcounter{pc}\newcounter{ppc}

%

\newenvironment{procedure}{
  \vspace{2mm}
  \setcounter{pc}{0}
  \begin{list}
    {\arabic{pc}.}{\usecounter{pc}
                   \setlength{\leftmargin}{8mm}
                   \setlength{\topsep}{0mm}
                   \setlength{\itemsep}{1mm}
                   \setlength{\parsep}{0mm}}}{
  \end{list}
  \vspace{2mm} }

\newenvironment{subprocedure}{
  \vspace{0.5mm}
  \setcounter{ppc}{0}
  \begin{list}
    {\alph{ppc}.}{\usecounter{ppc}
                  \setlength{\leftmargin}{\lmc}
                  \setlength{\topsep}{0mm}
                  \setlength{\itemsep}{0mm}
                  \setlength{\parsep}{0mm}}}{
  \end{list}
  \vspace{0.5mm} }

%

\newenvironment{text}{
  \vspace{2mm}
  \begin{list}
    {}{\setlength{\topsep}{0mm}
       \setlength{\leftmargin}{\lma}
       \setlength{\itemsep}{1mm}
       \setlength{\parsep}{0mm}}}{
  \end{list}
  \vspace{2mm} }

%

\newenvironment{definition}{
  \vspace{2mm}
  \begin{list}
    {}{\setlength{\topsep}{0mm}
       \setlength{\leftmargin}{\lmb}
       \setlength{\itemsep}{1mm}
       \setlength{\parsep}{0mm}}}{
  \end{list}
  \vspace{2mm} }

%

\begin{abstract}
This paper proposes a method for measuring semantic similarity between
words as a new tool for text analysis.  The similarity is measured on a
semantic network constructed systematically from a subset of the English
dictionary, LDOCE ({\it Longman Dictionary of Contemporary English\/}).
Spreading activation on the network can directly compute the similarity
between any two words in the {\it Longman Defining Vocabulary}, and
indirectly the similarity of all the other words in LDOCE.  The
similarity represents the strength of lexical cohesion or semantic
relation, and also provides valuable information about similarity and
coherence of texts.
\end{abstract}

%

\section{Introduction}

A text is not just a sequence of words, but it also has coherent
structure.  The meaning of each word in a text depends on the structure
of the text.  Recognizing the structure of text is an essential task in
text understanding.\cite{GroszS:86}

One of the valuable indicators of the structure of text is lexical
cohesion.\cite{HallidayH:76} \ Lexical cohesion is the relationship
between words, classified as follows:
\begin{procedure}
\item Reiteration:\\
  Molly likes {\it cats}. \ She keeps a {\it cat}.
\item Semantic relation:
  \begin{subprocedure}
  \item Desmond saw a {\it cat}. \ It was Molly's {\it pet}.
  \item Molly goes to the {\it north}. \ Not {\it east}.
  \item Desmond goes to a {\it theatre}. \ He likes {\it films}.
  \end{subprocedure}
\end{procedure}
Reiteration of words is easy to capture by morphological analysis. 
Semantic relation between words, which is the focus of this paper, is
hard to recognize by computers.

We consider lexical cohesion as semantic similarity between words. 
Similarity is computed by spreading activation (or association)
\cite{WaltzP:85} on a semantic network constructed systematically from
an English dictionary.  Whereas it is edited by some lexicographers, a
dictionary is a set of associative relation shared by the people in a
linguistic community.

The similarity between words is a mapping $\sigma$: $L \!\times\! L
\!\rightarrow\! [0,1]$, where $L$ is a set of words (or lexicon).  The 
following examples suggest the feature of the similarity:
\begin{tabbing}
\hspace{5mm}\=\hspace{22mm}\=\hspace{5mm}\=\hspace{18mm}\=\kill
\> $\sigma$({\tt cat}, \ {\tt pet})\>=\>0.133722 \> (similar),\\
\> $\sigma$({\tt cat}, \ {\tt mat})\>=\>0.002692 \> (dissimilar).
\end{tabbing}
The value of $\sigma(w,w')$ increases with strength of semantic relation
between $w$ and $w'$.

The following section examines related work in order to clarify the
nature of the semantic similarity.  Section~3 describes how the semantic
network is systematically constructed from the English dictionary.
Section~4 explains how to measure the similarity by spreading activation
on the semantic network.  Section~5 shows applications of the similarity
measure --- computing similarity between texts, and measuring coherence
of a text.  Section~6 discusses the theoretical aspects of the
similarity.

%

\section{Related Work on Measuring Similarity}

Words in a language are organized by two kinds of relationship.  One is
a syntagmatic relation: how the words are arranged in sequential texts.
The other is a paradigmatic relation: how the words are associated with
each other.  Similarity between words can be defined by either a
syntagmatic or a paradigmatic relation.

Syntagmatic similarity is based on co-occurrence data extracted from
corpora \cite{ChurchH:90}, definitions in dictionaries \cite{Wilks+:89},
and so on.  Paradigmatic similarity is based on association data
extracted from thesauri \cite{MorrisH:91}, psychological experiments
\cite{Osgood:52}, and so on.

This paper concentrates on paradigmatic similarity, because a
paradigmatic relation can be established both inside a sentence and
across sentence boundaries, while syntagmatic relations can be seen
mainly inside a sentence --- like syntax deals with sentence structure.
The rest of this section focuses on two related works on measuring
paradigmatic similarity --- a psycholinguistic approach and a
thesaurus-based approach.

\subsection{A Psycholinguistic Approach}

Psycholinguists have been proposed methods for measuring similarity.
One of the pioneering works is `semantic differential' \cite{Osgood:52}
which analyses meaning of words into a range of different dimensions
with the opposed adjectives at both ends (see Figure~1), and locates the
words in the semantic space.

\setlength{\unitlength}{1mm}
\begin{figure}[tb]\begin{center}\small
\begin{picture}(75,47)
\multiputlist( 37.5, 45.0)(0,0){\normalsize\bf ``polite''}
\thinlines
\matrixput( 20.0, 4.0)(5,0){8}(0,4){10}{\line(0,1){1}\line(0,-1){1}}
\matrixput( 20.0, 4.0)(0,4){10}(0,0){1}{\line(1,0){35}}
\multiputlist(16.0, 39.2)(0,-4)[rb]
{{\tt angular},{\tt weak},{\tt rough},{\tt active},{\tt small}, 
{\tt cold},{\tt good},{\tt tense},{\tt wet},{\tt fresh}}
\multiputlist(59.0, 39.2)(0,-4)[lb]
{{\tt rounded},{\tt strong},{\tt smooth},{\tt passive},{\tt large}
,{\tt hot},{\tt bad},{\tt relaxed},{\tt dry},{\tt stale}}
\thicklines
\begin{drawjoin}
  \jput(40.0, 40){\makebox(0,0){$\bullet$}}   
  \jput(45.0, 36){\makebox(0,0){$\bullet$}}   
  \jput(49.0, 32){\makebox(0,0){$\bullet$}}   
  \jput(32.0, 28){\makebox(0,0){$\bullet$}}   
  \jput(37.0, 24){\makebox(0,0){$\bullet$}}   
  \jput(37.0, 20){\makebox(0,0){$\bullet$}}   
  \jput(25.0, 16){\makebox(0,0){$\bullet$}}   
  \jput(47.0, 12){\makebox(0,0){$\bullet$}}   
  \jput(37.0,  8){\makebox(0,0){$\bullet$}}   
  \jput(31.0,  4){\makebox(0,0){$\bullet$}}   
\end{drawjoin}
\end{picture}\\
\small
{\bf Figure  1.} \ A psycholinguistic measurement \\
(semantic differential \cite{Osgood:52}).
\end{center}\end{figure}

\begin{figure*}[tb]\begin{center}
\hspace*{4mm}
\begin{minipage}{65mm}\small\setlength{\baselineskip}{3.5mm}
  \hspace*{-2mm}
    {\bf red$^{\mbox{\tiny\bf 1}}$} /red/ {\it adj} {\bf -dd-}
  \ {\bf 1} of the colour of blood or fire:
    {\it a red rose/dress} $|$ {\it We painted the door red.}
    --- see also {\bf like a red rag to a bull} ({\sc rag}$^{1}$)
  \ {\bf 2} (of human hair) of a bright brownish orange or copper colour
  \ {\bf 3} (of the human skin) pink, usu. for a short time:
    {\it I turned red with embarrassment/anger.} $|$
    {\it The child's eye} (= the skin round the eyes)
    {\it were red from crying.}
  \ {\bf 4} (of wine) of a dark pink to dark purple colour
  --- $\sim$ness {\it n} [U]
\end{minipage}
\hspace{8mm}
\begin{minipage}{83mm}\small\setlength{\baselineskip}{3.25mm}
\begin{verbatim}
(red adj                   ; headword, word-class
  ((of the colour)         ; unit 1 -- head-part
   (of blood or fire) )    ;           det-part
  ((of a bright brownish  orange or copper colour)
   (of human hair) )
  (pink                    ; unit 3 -- head-part
   (usu for a short time)  ;           det-part 1
   (of the human skin) )   ;           det-part 2
  ((of a dark pink to dark purple colour)
   (of wine) ))
\end{verbatim}
\end{minipage}\\
\vspace{3.5mm}\small
{\bf Figure  2.} \ A sample entry of LDOCE \ and \ 
    a corresponding entry of {\it Glosseme\/} (in S-expression).
\end{center}\end{figure*}

\begin{figure*}[tb]\begin{center}
\begin{minipage}{164mm}\footnotesize\setlength{\baselineskip}{3.0mm}
\begin{verbatim}
  (red_1  (adj)  0.000000    ;;   headword, word-class, and activity-value
    ;;   referant
    (+ ;;   subreferant 1
       (0.333333  ;;   weight of subreferant 1
         (* (0.001594      of_1) (0.001733     the_1) (0.001733     the_2) (0.042108  colour_1) 
            (0.042108  colour_2) (0.000797      of_1) (0.539281   blood_1) (0.000529      or_1) 
            (0.185058    fire_1) (0.185058    fire_2) ))
       ;;   subreferant 2
       (0.277778
         (* (0.000278      of_1) (0.000196       a_1) (0.030997  bright_1) (0.065587   brown_1)
            (0.466411  orange_1) (0.000184      or_1) (0.385443  copper_1) (0.007330  colour_1)
            (0.007330  colour_2) (0.000139      of_1) (0.009868   human_1) (0.009868   human_2) 
            (0.016372    hair_1) ))
       ;;   subreferant 3
       (0.222222
         (* (0.410692    pink_1) (0.410692    pink_2) (0.003210     for_1) (0.000386       a_1) 
            (0.028846   short_1) (0.006263    time_1) (0.000547      of_1) (0.000595     the_1) 
            (0.000595     the_2) (0.038896   human_1) (0.038896   human_2) (0.060383    skin_1) ))
       ;;   subreferant 4
       (0.166667
         (* (0.000328      of_1) (0.000232       a_1) (0.028368    dark_1) (0.028368    dark_2) 
            (0.123290    pink_1) (0.123290    pink_2) (0.000273      to_1) (0.000273      to_2) 
            (0.000273      to_3) (0.028368    dark_1) (0.028368    dark_2) (0.141273  purple_1) 
            (0.141273  purple_2) (0.008673  colour_1) (0.008673  colour_2) (0.000164      of_1) 
            (0.338512    wine_1) )))
    ;;   refere
    (* (0.031058   apple_1) (0.029261   blood_1) (0.008678  colour_1) (0.009256    comb_1) 
       (0.029140  copper_1) (0.009537 diamond_1) (0.003015    fire_1) (0.073762   flame_1) 
       (0.005464     fox_1) (0.005152   heart_1) (0.098349    lake_2) (0.007025     lip_1) 
       (0.029140  orange_1) (0.007714  pepper_1) (0.196698    pink_1) (0.012294    pink_2) 
       (0.098349    pink_2) (0.018733  purple_2) (0.028100  purple_2) (0.098349     red_2) 
       (0.196698     red_2) (0.004230  signal_1) ))
\end{verbatim}\end{minipage}\\
\vspace{3.5mm}\small
{\bf Figure 3.} \ A sample node of {\it Paradigme\/} (in S-expression).
\end{center}\end{figure*}

Recent works on knowledge representation are somewhat related to
Osgood's semantic differential.  Most of them describe meaning of words
using special symbols like microfeatures \cite{WaltzP:85,Hendler:89}
that correspond to the semantic dimensions.

However, the following problems arise from the semantic differential
procedure as measurement of meaning.  The procedure is not based on the
denotative meaning of a word, but only on the connotative emotions
attached to the word; \ it is difficult to choose the relevant
dimensions, i.e.~the dimensions required for the sufficient semantic
space.

\subsection{A Thesaurus-based Approach}

Morris and Hirst \nocite{MorrisH:91}[1991] used Roget's thesaurus as
knowledge base for determining whether or not two words are semantically
related.  For example, the semantic relation of {\tt truck}/{\tt car}
and {\tt drive}/{\tt car} are captured in the following way:
\begin{procedure}
\item {\tt truck} $\!\in\!$ {\it vehicle\/}
                  $\!\ni\!$ {\tt car} \\
      (both are included in the {\it vehicle\/} class),
\item {\tt drive} $\in$ {\it journey\/} $\rightarrow$
                        {\it vehicle\/} $\ni$ {\tt car} \\
      ({\it journey\/} refers to {\it vehicle\/}).
\end{procedure}

This method can capture almost all types of semantic relations (except
emotional and situational relation), such as paraphrasing by
superordinate (ex. {\tt cat}/{\tt pet}), systematic relation (ex. {\tt
north}/{\tt east}), and non-systematic relation (ex. {\tt theatre}/{\tt
film}).

However, thesauri provide neither information about semantic difference
between words juxtaposed in a category, nor about strength of the
semantic relation between words --- both are to be dealt in this paper.
The reason is that thesauri are designed to help writers find relevant
words, not to provide the meaning of words.

%

\section{Paradigme: A Field for Measuring Similarity}

We analyse word meaning in terms of the semantic space defined by a
semantic network, called {\it Paradigme\/}.  {\it Paradigme\/} is
systematically constructed from {\it Gloss\`eme\/}, a subset of an
English dictionary.

\subsection{Gloss\`eme --- A Closed Subsystem of English}

A dictionary is a closed paraphrasing system of natural language.  Each
of its headwords is defined by a phrase which is composed of the
headwords and their derivations.  A dictionary, viewed as a whole, looks
like a tangled network of words.

We adopted {\it Longman Dictionary of Contemporary English\/} (LDOCE)
\nocite{LDOCE}[1987] as such a closed system of English.  LDOCE has a
unique feature that each of its $56{,}000$ headwords is defined by using 
the words in {\it Longman Defining Vocabulary\/} (hereafter, LDV) and 
their derivations.  LDV consists of $2{,}851$ words (as the headwords in 
LDOCE) based on the survey of restricted vocabulary \cite{West:53}.

We made a reduced version of LDOCE, called {\it Gloss\`eme\/}.  {\it
Gloss\`eme\/} has every entry of LDOCE whose headword is included in
LDV.  Thus, LDV is defined by {\it Gloss\`eme\/}, and {\it Gloss\`eme\/}
is composed of LDV.  {\it Gloss\`eme\/} is a closed subsystem of
English.

{\it Gloss\`eme\/} has $2{,}851$ entries that consist of $101{,}861$ 
words (35.73 words/entry on the average).  An item of {\it Gloss\`eme\/} 
has a {\sl headword\/}, a {\sl word-class\/}, and one or more {\sl 
unit\/}s corresponding to numbered definitions in the entry of LDOCE.  
Each unit has one {\sl head-part\/} and several {\sl det-part\/}s.  The 
head-part is the first phrase in the definition, which describes the 
broader meaning of the headword.  The det-parts restrict the meaning of 
the head-part.  (See Figure~2.)

\subsection{Paradigme --- A Semantic Network}

We then translated {\it Gloss\`eme} into a semantic network {\it
Paradigme\/}.  Each entry in {\it Gloss\`eme} is mapped onto a node in
{\it Paradigme\/}.  {\it Paradigme\/} has $2{,}851$ nodes and $295{,}914$
unnamed links between the nodes (103.79 links/node on the average).
Figure~3 shows a sample node {\tt red\_1}.  Each node consists of a {\sl
headword\/}, a {\sl word-class\/}, an {\sl activity-value\/}, and two
sets of {\sl link\/}s: a {\sl r\'ef\'erant\/} and a {\sl
r\'ef\'er\'e\/}.

A r\'ef\'erant of a node consists of several {\sl subr\'ef\'erant\/}s
correspond to the units of {\it Gloss\`eme\/}.  As shown in Figure~2 
and~3, a morphological analysis maps the word {\tt brownish} in the 
second unit onto a link to the node {\tt brown\_1}, and the word {\tt 
colour} onto two links to {\tt colour\_1} (adjective) and {\tt 
colour\_2} (noun).

A r\'ef\'er\'e of a node $p$ records the nodes referring to $p$.  For
example, the r\'ef\'er\'e of {\tt red\_1} is a set of links to nodes
(ex. {\tt apple\_1}) that have a link to {\tt red\_1} in their
r\'ef\'erants. The r\'ef\'er\'e provides information about the {\it
extension\/} of {\tt red\_1}, not the {\it intension\/} shown in the
r\'ef\'erant.

\begin{figure*}[tb]\begin{center}
\begin{picture}(120,27)
  \thicklines\put( 15,7.5){\oval(30, 15)}
    \put( 12, 7.5){\oval( 3,1.5)}
    \put( 18, 7.5){\oval( 3,1.5)}
    \put( 12,22.5){\vector(0,-1){15}}
    \put( 12,  25){\makebox(0,0){$w$}}
    \put(7.5,  19){\makebox(0,0){$s(w)$}}
  \thicklines\put( 60,7.5){\oval(30, 15)}
    \put( 57, 7.5){\oval( 3,1.5)}
    \put( 63, 7.5){\oval( 3,1.5)}
    \put( 57,22.5){\vector(0,-1){15}}
    \put( 57,  25){\makebox(0,0){$w$}}
    \put(52.5, 19){\makebox(0,0){$s(w)$}}
  \thinlines \put( 57,7.5){\oval( 6,   3)}
             \put( 57,7.5){\oval(10,   5)}
             \put( 57,7.5){\oval(15, 7.5)}
             \put( 57,7.5){\oval(21,10.5)}
  \thicklines\put(105,7.5){\oval(30, 15)}
    \put(102,7.5){\oval( 3,1.5)}
    \put(108,7.5){\oval( 3,1.5)}
    \put(108,7.5){\vector(0, 1){15}}
    \put(108, 25){\makebox(0,0){$w'$}}
    \put(113, 19){\makebox(0,0){$s(w')$}}
  \thinlines \put(102,7.5){\oval( 6,   3)}
             \put(102,7.5){\oval(10,   5)}
             \put(102,7.5){\oval(15, 7.5)}
             \put(102,7.5){\oval(21,10.5)}
\end{picture}\\
\vspace{3mm}\small
{\bf Figure  4.} \ Process of measuring the similarity $\sigma(w,w')$
                   on {\it Paradigme\/}.\\
                   (1) Start activating $w$.
               \ \ (2) Produce an activated pattern.
               \ \ (3) Observe activity of $w'$.
\end{center}\end{figure*}

Each link has {\sl thickness\/} $t_k$, which is computed from the
frequency of the word $w_k$ in {\it Gloss\`eme\/} and other information,
and normalized as $\sum t_k \!=\! 1$ in each subr\'ef\'erant or
r\'ef\'er\'e.  Each subr\'ef\'erant also has {\sl thickness\/} (for
example, 0.333333 in the first subr\'ef\'erant of {\tt red\_1}), which
is computed by the order of the units which represents significance of
the definitions.  Appendix A describes the structure of {\it
Paradigme\/} in detail.

%

\begin{figure}[tb]\begin{center}
\\
\vspace{-1mm}\small
{\bf Figure 5.} \ An activated pattern produced from {\tt red}\\ 
(changing of activity values of 10 nodes \\
holding highest activity at $T\!=\!10$).
\end{center}\end{figure}

\section{Computing Similarity between Words}

Similarity between words is computed by spreading activation on {\it
Paradigme\/}.  Each of its nodes can hold activity, and it moves through
the links.  Each node computes its activity value $v_i(T\!+\!1)$ at time
$T\!+\!1$ as follows:
\[
  v(T\!+\!1) = \phi\left(R_i(T), R'_i(T), e_i(T)\right),
\]
where $R_i(T)$ and $R'_i(T)$ are the sum of weighted activity (at time
$T$) of the nodes referred in the r\'ef\'erant and r\'ef\'er\'e
respectively.  And, $e_i(T)$ is activity given from outside (at time
$T$); \ to `activate a node' is to let $e_i(T) \!>\! 0$.  The output
function $\phi$ sums up three activity values in appropriate proportion
and limits the output value to [0,1].  Appendix B gives the details of
the spreading activation.

\subsection{Measuring Similarity}

Activating a node for a certain period of time causes the activity to
spread over {\it Paradigme} and produce an activated pattern on it.  The
activated pattern approximately gets equilibrium after 10 steps, whereas
it will never reach the actual equilibrium.  The pattern thus produced
represents the meaning of the node or of the words related to the node
by morphological analysis\footnote{The morphological analysis maps all
words derived by 48 affixes in LDV onto their root forms (i.e.~headwords
of LDOCE).}.

The activated pattern, produced from a word $w$, suggests similarity
between $w$ and any headword in LDV.  The similarity $\sigma(w,w')
\!\in\! [0,1]$ is computed in the following way.  (See also Figure~4.)
\begin{procedure}
\item Reset activity of all nodes in {\it Paradigme\/}.
\item Activate $w$ with strength $s(w)$ for 10 steps, 
      where $s(w)$ is significance of the word $w$. \\
      Then, an activated pattern $P(w)$ is produced 
      on {\it Paradigme\/}.
\item Observe $a(P(w),w')$ --- an activity value 
      of the node $w'$ in $P(w)$. \\
      Then, $\sigma(w,w')$ is $s(w')\!\cdot\!a(P(w),w')$.
\end{procedure}

The word significance $s(w) \!\in\! [0,1]$ is defined as the normalized
information of the word $w$ in the corpus \cite{West:53}.  For example,
the word {\tt red} appears $2{,}308$ times in the 5,487,056-word corpus, 
and the word {\tt and} appears $106{,}064$ times. So, $s({\tt red})$ and 
$s({\tt and})$ are computed as follows:
\[   s({\tt red})
     = \vbox{\hbox{\Large$\frac{\ -\log(2308/5487056)\ }{-\log(1/5487056)}$}}
     = 0.500955 \ , \]
\[ s({\tt and})
     = \vbox{\hbox{\Large$\frac{-\log(106064/5487056)}{-\log(1/5487056)}$}}
     = 0.254294 \ . \]
We estimated the significance of the words excluded from the word list 
\cite{West:53} at the average significance of their word classes.  This 
interpolation virtually enlarged West's $5{,}000{,}000$-word corpus.

For example, let us consider the similarity between {\tt red} and {\tt
orange}.  First, we produce an activated pattern $P(\mbox{\tt red})$ on
{\it Paradigme\/}.  (See Figure~5.) \ In this case, both of the nodes
{\tt red\_1} (adjective) and {\tt red\_2} (noun) are activated with
strength $s({\tt red}) \!=\! 0.500955$.  Next, we compute $s(\mbox{\tt
orange}) \!=\! 0.676253$, and observe $a(P(\mbox{\tt red}), \mbox{\tt
orange}) \!=\! 0.390774$. Then, the similarity between {\tt red} and
{\tt orange} is obtained as follows:
\begin{tabbing}
  \hspace{5mm}
  $\sigma(\mbox{\tt red},\mbox{\tt orange})$
    \= = \= $0.676253 \cdot 0.390774$ \\
    \> = \> 0.264262 .
\end{tabbing}

\subsection{Examples of Similarity between Words}

The procedure described above can compute the similarity $\sigma(w,w')$
between any two words $w,w'$ in LDV and their derivations.  Computer
programs of this procedure --- spreading activation (in C),
morphological analysis and others (in Common Lisp) --- can compute
$\sigma(w,w')$ within 2.5 seconds on a workstation (SPARCstation 2).

The similarity $\sigma$ between words works as an indicator of the
lexical cohesion.  The following examples illustrate that $\sigma$
increases with the strength of semantic relation:
\begin{tabbing}
\hspace{5mm}\=\hspace{22mm}\=\hspace{22mm}\=\hspace{5mm}\=\kill
\> $\sigma$({\tt wine},     \> {\tt alcohol})    \> = \> 0.118078 , \\
\> $\sigma$({\tt wine},     \> {\tt line})       \> = \> 0.002040 , \\
\> $\sigma$({\tt big},      \> {\tt large})      \> = \> 0.120587 , \\
\> $\sigma$({\tt clean},    \> {\tt large})      \> = \> 0.004943 , \\
\> $\sigma$({\tt buy},      \> {\tt sell})       \> = \> 0.135686 , \\
\> $\sigma$({\tt buy},      \> {\tt walk})       \> = \> 0.007993 .
\end{tabbing}
The similarity $\sigma$ also increases with the co-occurrence tendency
of words, for example:
\begin{tabbing}
\hspace{5mm}\=\hspace{22mm}\=\hspace{22mm}\=\hspace{5mm}\=\kill
\> $\sigma$({\tt waiter},   \> {\tt restaurant}) \> = \> 0.175699 , \\
\> $\sigma$({\tt computer}, \> {\tt restaurant}) \> = \> 0.003268 , \\
\> $\sigma$({\tt red},      \> {\tt blood})      \> = \> 0.111443 , \\
\> $\sigma$({\tt green},    \> {\tt blood})      \> = \> 0.002268 , \\
\> $\sigma$({\tt dig},      \> {\tt spade})      \> = \> 0.116200 , \\
\> $\sigma$({\tt fly},      \> {\tt spade})      \> = \> 0.003431 .
\end{tabbing}
Note that $\sigma(w,w')$ has direction (from $w$ to $w'$), so that
$\sigma(w,w')$ may not be equal to $\sigma(w',w)$:
\begin{tabbing}
\hspace{5mm}\=\hspace{22mm}\=\hspace{22mm}\=\hspace{5mm}\=\kill
\> $\sigma$({\tt films},    \> {\tt theatre})    \> = \> 0.178988 , \\
\> $\sigma$({\tt theatre},  \> {\tt films})      \> = \> 0.068927 .
\end{tabbing}

Meaningful words should have higher similarity; \ meaningless words
(especially, function words) should have lower similarity.  The
similarity $\sigma(w,w')$ increases with the significance $s(w)$ and
$s(w')$ that represent meaningfulness of $w$ and $w'$:
\begin{tabbing}
\hspace{5mm}\=\hspace{22mm}\=\hspace{22mm}\=\hspace{5mm}\=\kill
\> $\sigma$({\tt north},    \> {\tt east})       \> = \> 0.100482 , \\
\> $\sigma$({\tt to},       \> {\tt theatre})    \> = \> 0.007259 , \\
\> $\sigma$({\tt films},    \> {\tt of})         \> = \> 0.005914 , \\
\> $\sigma$({\tt to},       \> {\tt the})        \> = \> 0.002240 .
\end{tabbing}
Note that the reflective similarity $\sigma(w,w)$ also depends on the
significance $s(w)$, so that $\sigma(w,w) \leq 1$:
\begin{tabbing}
\hspace{5mm}\=\hspace{22mm}\=\hspace{22mm}\=\hspace{5mm}\=\kill
\> $\sigma$({\tt waiter},   \> {\tt waiter})     \> = \> 0.596803 , \\
\> $\sigma$({\tt of},       \> {\tt of})         \> = \> 0.045256 .
\end{tabbing}

\begin{figure}[tb]\begin{center}
\\
\vspace{5mm}\small
{\bf Figure 6.} \ Measuring similarity of entra words \\
                  as the similarity between word lists.
\end{center}\end{figure}

\begin{figure}[tb]\begin{center}
%
\\
\vspace{-1mm}\small
{\bf Figure 7.} \ An activated pattern produced from \\
the word list: \{{\tt red}, {\tt alcoholic}, {\tt drink}\}.
\end{center}\end{figure}

\subsection{Similarity of Extra Words}

The similarity of words in LDV and their derivations is measured
directly on {\it Paradigme\/}; \ the similarity of extra words is measured
indirectly on {\it Paradigme\/} by treating an extra word as a word list
$W \!=\! \{w_1, \!\cdots\!, w_n\}$ of its definition in LDOCE.  (Note
that each $w_i \in W$ is included in LDV or their derivations.)

The similarity between the word lists $W, W'$ is defined as follows. 
(See also Figure~6.)
\begin{text}
\item $\sigma(W,W') 
      = \psi\left(\sum_{w' \in W'} s(w')\!\cdot\!a(P(W),w')\right)$,
\end{text}
where $P(W)$ is the activated pattern produced from $W$ by activating
each $w_i\!\in\!W$ with strength $s(w_i)^2/\sum{s(w_k)}$ for 10 steps.
And, $\psi$ is an output function which limits the value to [0,1].

As shown in Figure~7, {\tt bottle\_1} and {\tt wine\_1} have high
activity in the pattern produced from the phrase ``red alcoholic
drink''.  So, we may say that the overlapped pattern implies ``a bottle
of wine''.

For example, the similarity between {\tt linguistics} and {\tt
stylistics}, both are the extra words, is computed as follows:
\begin{tabbing}
\hspace{5mm}\=\hspace{5mm}\=\hspace{7.5mm}\=\hspace{2mm}\=\kill
\> $\sigma$({\tt linguistics}, {\tt stylistics}) \\
\> \> = $\sigma$(
        \> \{ \> {\tt the}, {\tt study}, {\tt of}, {\tt language}, {\tt in}, \\
\> \>   \>    \> {\tt general}, {\tt and}, {\tt of}, {\tt particular}, \\
\> \>   \>    \> {\tt languages}, {\tt and}, {\tt their}, {\tt structure}, \\
\> \>   \>    \> {\tt and}, {\tt grammar}, {\tt and}, {\tt history}\}, \\
\> \>   \> \{ \> {\tt the}, {\tt study}, {\tt of}, {\tt style}, {\tt in},\\
\> \>   \>    \> {\tt written}, {\tt or}, {\tt spoken}, {\tt language}\} )\\
\> \> = \ 0.140089 .
\end{tabbing}

Obviously, both $\sigma(W,w)$ and $\sigma(w,W)$, where $W$ is an extra
word and $w$ is not, are also computable.  Therefore, we can compute the
similarity between any two headwords in LDOCE and their derivations.

%

\section{Applications of the Similarity}

This section shows the application of the similarity between words to
text analysis --- measuring similarity between texts, and measuring text
coherence.

\subsection{Measuring Similarity between Texts}

Suppose a text is a word list without syntactic structure.  Then, the
similarity $\sigma(X,X')$ between two texts $X,X'$ can be computed as
the similarity of extra words described above.

\begin{figure}[tb]\begin{center}\small
\begin{picture}(60,49)(0,7)
  \thicklines
  \put(30, 35){\oval(60, 26)}
  \put(30, 41){\oval(24, 10)}
  \put(30, 29){\oval(24, 10)}
  \put(18, 33){\oval(24, 10)}
  \put(42, 37){\oval(24, 10)}
  \thinlines
  \put(30, 41){\oval( 19,  8)}
  \put(30, 41){\oval( 14,  6)}
  \put(30, 41){\oval(  9,  4)}
  \put(30, 41){\oval(  5,  2)}
  \drawline(20, 45)(30, 53)(40,  45)
  \drawline(6.5,31)(18, 17)(29.5,31)
  \drawline(20, 25)(30, 17)(40,  25)
  \drawline(31, 34)(42, 17)(53,  34)
  \put(24, 55){\makebox(0,0){text:}}
  \put(30, 55){\makebox(0,0){$X$}}
  \put(18, 14.5){\makebox(0,0){$X'_1$}}
  \put(30, 14.5){\makebox(0,0){$X'_2$}}
  \put(42, 14.5){\makebox(0,0){$X'_3$}}
  \put(30, 11.5){\makebox(0,0){$\underbrace{\hspace{30mm}}$}}
  \put(30,  8.5){\makebox(0,0){episodes}}
\end{picture}\\
\vspace{1mm}\small
{\bf Figure 8.} \ Episode association on {\it Paradigme\/}\\
(recalling the most similar episode in memory).
\end{center}\end{figure}

The following examples suggest that the similarity between texts
indicates the strength of coherence relation between them:
\begin{tabbing}
\hspace{5mm}\=\hspace{3.5mm}\=\hspace{48mm}\=\hspace{5mm}\=\kill
\> $\sigma$( \> {\tt "I have a hammer."}, \\
\>           \> {\tt "Take some nails."} )       \> = \> 0.100611 ,\\
\> $\sigma$( \> {\tt "I have a hammer."}, \\
\>           \> {\tt "Take some apples."} )      \> = \> 0.005295 , \\
\> $\sigma$( \> {\tt "I have a pen."}, \\
\>           \> {\tt "Where is ink?"} )          \> = \> 0.113140 , \\
\> $\sigma$( \> {\tt "I have a pen."}, \\
\>           \> {\tt "Where do you live?"} )     \> = \> 0.007676 .
\end{tabbing}

It is worth noting that meaningless iteration of words (especially, of
function words) has less influence on the text similarity:
\begin{tabbing}
\hspace{5mm}\=\hspace{3.5mm}\=\hspace{48mm}\=\hspace{5mm}\=\kill
\> $\sigma$( \> {\tt "It is a dog."}, \\
\>           \> {\tt "That must be your dog."} ) \> = \> 0.252536 , \\
\> $\sigma$( \> {\tt "It is a dog."}, \\
\>           \> {\tt "It is a log."} )           \> = \> 0.053261 .
\end{tabbing}

The text similarity provides a semantic space for text retrieval --- to
recall the most similar text in $\{X'_1, \cdots, X'_n\}$ to the given
text $X$.  Once the activated pattern $P(X)$ of the text $X$ is produced
on {\it Paradigme\/}, we can compute and compare the similarity
$\sigma(X,X'_1), \cdots, \sigma(X,X'_n)$ immediately.  (See Figure~8.)

\subsection{Measuring Text Coherence}

Let us consider the reflective similarity $\sigma(X,X)$ of a text $X$,
and use the notation $c(X)$ for $\sigma(X,X)$.  Then, $c(X)$ can be
computed as follows:
\begin{text}
\item $c(X) = \psi\left(\sum_{w \in X} s(w) a(P(X), w)\right)$.
\end{text}
The activated pattern $P(X)$, as shown in Figure~7, represents the
average meaning of $w_i \!\in \!X$.  So, $c(X)$ represents cohesiveness
of $X$ --- or semantic closeness of $w \!\in\! X$, or semantic
compactness of $X$.  (It is also closely related to {\it distortion\/}
in clustering.)

The following examples suggest that $c(X)$ indicates the strength of
coherence of $X$:
\begin{text}
\item $c$ (\begin{minipage}[t]{70mm}
            {\tt "She opened the world with her \\
               \  typewriter.  Her work was typing. \\
               \  But She did not type quickly."} )
           \end{minipage} \\ \hspace*{2mm} = 0.502510 
           \ \ (coherent),
\item $c$ (\begin{minipage}[t]{70mm}
            {\tt "Put on your clothes at once. \\
               \  I can not walk ten miles. \\
               \  There is no one here but me."} )
           \end{minipage} \\ \hspace*{2mm} = 0.250840
           \ \ (incoherent).
\end{text}

However, a cohesive text can be incoherent; \ the following example
shows cohesiveness of the incoherent text --- three sentences randomly
selected from LDOCE:
\begin{text}
\item $c$ (\begin{minipage}[t]{70mm}
            {\tt "I saw a lion. \\
               \  A lion belongs to the cat family. \\
               \  My family keeps a pet."} )
           \end{minipage} \\ \hspace*{2mm} = 0.560172
           \ \ (incoherent, but cohesive).
\end{text}
Thus, $c(X)$ can not capture all the aspects of text coherence.  This is
because $c(X)$ is based only on the lexical cohesion of the words in
$X$.

%

\section{Discussion}

The structure of {\it Paradigme\/} represents the knowledge system of
English, and an activated state produced on it represents word meaning.
This section discusses the nature of the structure and states of {\it
Paradigme\/}, and also the nature of the similarity computed on it.

\subsection{Paradigme and Semantic Space}

The set of all the possible activated patterns produced on {\it
Paradigme\/} can be considered as a semantic space where each state is
represented as a point.  The semantic space is a $2{,}851$-dimensional
hypercube; \ each of its edges corresponds to a word in LDV.

LDV is selected according to the following information: the word
frequency in written English, and the range of contexts in which each
word appears.  So, LDV has a potential for covering all the concepts
commonly found in the world.

This implies the completeness of LDV as dimensions of the semantic
space.  Osgood's semantic differential procedure \nocite{Osgood:52} 
[1952] used 50 adjective dimensions; \ our semantic measurement uses 
$2{,}851$ dimensions with completeness and objectivity.

Our method can be applied to construct a semantic network from an
ordinary dictionary whose defining vocabulary is not restricted.  Such a
network, however, is too large to spread activity over it.  {\it
Paradigme\/} is the small and complete network for measuring the
similarity.

\subsection{Connotation and Extension of Words}

The proposed similarity is based only on the denotational and
intensional definitions in the dictionary LDOCE.  Lack of the
connotational and extensional knowledge causes some unexpected results
of measuring the similarity.  For example, consider the following
similarity:
\begin{tabbing}
\hspace{5mm}\=\hspace{14mm}\=\hspace{12mm}\=\hspace{5mm}\=\kill
\> $\sigma$({\tt tree},     \> {\tt leaf})    \> = \> 0.008693 .
\end{tabbing}

This is due to the nature of the dictionary definitions --- they only
indicate sufficient conditions of the headword.  For example, the
definition of {\tt tree} in LDOCE tells nothing about leaves:
\begin{definition}
\item \hspace*{-3mm} {\bf tree} \ {\it n\/}
      \ {\bf 1} \ a tall plant with a wooden trunk and branches, that 
      lives for many years
      \ {\bf 2} \ a bush or other plant with a treelike form
      \ {\bf 3} \ a drawing with a branching form, esp.~as used for 
      showing family relationships
\end{definition}
However, the definition is followed by pictures of leafy trees providing
readers with connotational and extensional stereotypes of trees.

\subsection{Paradigmatic and Syntagmatic Similarity}

In the proposed method, the definitions in LDOCE are treated as word
lists, though they are phrases with syntactic structures.  Let us
consider the following definition of {\tt lift}:
\begin{definition}
\item \hspace*{-3mm} {\bf lift} \ {\it v\/}
      \ {\bf 1} \ to bring from a lower to a higher level; \ raise
      \ {\bf 2} \ (of movable parts) to be able to be lifted 
      \ {\bf 3} \ $\cdots$
\end{definition}
Anyone can imagine that something is moving upward.  But, such a
movement can not be represented in the activated pattern produced from
the phrase.  The meaning of a phrase, sentence, or text should be
represented as pattern changing in time, though what we need is static
and paradigmatic relation.

This paradox also arises in measuring the similarity between texts and
the text coherence.  As we have seen in Section~5, there is a difference
between the similarity of texts and the similarity of word lists, and
also between the coherence of a text and cohesiveness of a word list.

However, so far as the similarity between words is concerned, we assume 
that activated patterns on {\it Paradigme\/} will approximate the 
meaning of words, like a still picture can express a story.

%

\section{Conclusion}

We described measurement of semantic similarity between words.  The
similarity between words is computed by spreading activation on the
semantic network {\it Paradigme\/} which is systematically constructed
from a subset of the English dictionary LDOCE.  {\it Paradigme\/} can
directly compute the similarity between any two words in LDV, and
indirectly the similarity of all the other words in LDOCE.

The similarity between words provides a new method for analysing the
structure of text.  It can be applied to computing the similarity
between texts, and measuring the cohesiveness of a text which suggests
coherence of the text, as we have seen in Section~5.  And, we are now
applying it to text segmentation \cite{GroszS:86,Youmans:91}, i.e.~to
capture the shifts of coherent scenes in a story.

In future research, we intend to deal with syntagmatic relations between
words.  Meaning of a text lies in the texture of paradigmatic and
syntagmatic relations between words \cite{Hjelmslev:43}.  {\it
Paradigme\/} provides the former dimension --- an associative system of
words --- as a screen onto which the meaning of a word is projected like
a still picture.  The latter dimension --- syntactic process --- will be
treated as a film projected dynamically onto {\it Paradigme\/}.  This
enables us to measure the similarity between texts as a syntactic
process, not as word lists.

We regard {\it Paradigme\/} as a field for the interaction between text
and episodes in memory --- the interaction between what one is hearing
or reading and what one knows \cite{Schank:90}.  The meaning of words,
sentences, or even texts can be projected in a uniform way on {\it
Paradigme\/}, as we have seen in Section~4 and~5.  Similarly, we can
project text and episodes, and recall the most relevant episode for
interpretation of the text.

%

\section*{Appendix A. \ Structure of Paradigme\\
--- Mapping Gloss\`eme onto Paradigme}

The semantic network {\it Paradigme\/} is systematically constructed
from the small and closed English dictionary {\it Gloss\`eme}.  Each
entry of {\it Gloss\`eme\/} is mapped onto a node of {\it Paradigme\/}
in the following way.  (See also Figure~2 and~3.)

{\bf Step 1.}\ For each entry $G_i$ in {\it Gloss\`eme\/}, map each 
unit $u_{ij}$ in $G_i$ onto a subr\'ef\'erant $s_{ij}$ of the 
corresponding node $P_i$ in {\it Paradigme\/}.  Each word $w_{ijn} 
\!\in\! u_{ij}$ is mapped onto a link or links in $s_{ij}$, in the 
following way:
\begin{procedure}
\item Let $t_{n}$ be the reciprocal of the number of appearance of 
      $w_{ijn}$ (as its root form) in {\it Gloss\`eme}.
\item If $w_{ijn}$ is in a head-part, let $t_{n}$ be doubled.
\item Find nodes $\{p_{n1}, p_{n2}, \cdots \}$ corresponds to $w_{ijn}$ 
      (ex.~{\tt red} $\!\rightarrow\!$ \{{\tt red\_1}, {\tt red\_2}\}).
      Then, divide $t_{n}$ into $\{t_{n1}, t_{n2}, \cdots \}$ in 
      proportion to their frequency.
\item Add links $l_{n1}, l_{n2}, \cdots$ to $s_{ij}$, where $l_{nm}$ is 
      a link to the node $p_{nm}$ with thickness $t_{nm}$.
\end{procedure}
Thus, $s_{ij}$ becomes a set of links: $\{l_{ij1}, l_{ij2}, \cdots \}$,
where $l_{ijk}$ is a link with thickness $t_{ijk}$.  Then, normalize
thickness of the links as $\sum_k t_{ijk} \!=\! 1$, in each $s_{ij}$.

{\bf Step 2.}\ For each node $P_i$, compute thickness $h_{ij}$ of
each subr\'ef\'erant $s_{ij}$ in the following way:
\begin{procedure}
\item Let $m_i$ be the number of subr\'ef\'erants of $P_i$.
\item Let $h_{ij}$ be $2m_i \!-\! 1 \!-\! j$.\\
      (Note that $h_{i1}:h_{im}$ = $2:1$.)
\item Normalize thickness $h_{ij}$ as $\sum_j h_{ij} \!=\! 1$, in each 
      $P_i$.
\end{procedure}

{\bf Step 3.}\ Generate r\'ef\'er\'e of each node in {\it Paradigme\/}, 
in the following way:
\begin{procedure}
\item For each node $P_i$ in {\it Paradigme\/}, let its r\'ef\'er\'e 
      $r_i$ be an empty set.
\item For each $P_i$, for each subr\'ef\'erant $s_{ij}$ of $P_i$, 
      for each link $l_{ijk}$ in $s_{ij}$:
      \begin{subprocedure}
      \item Let $p_{ijk}$ be the node referred by $l_{ijk}$, and let 
            $t_{ijk}$ be thickness of $l_{ijk}$.
      \item Add a new link $l'$ to r\'ef\'er\'e of $p_{ijk}$, where $l'$ 
            is a link to $P_i$ with thickness $t' = h_{ij} \!\cdot\! 
            t_{ijk}$.
      \end{subprocedure}
\item Thus, each $r_i$ becomes a set of links: $\{l'_{i1}, l'_{i2}, 
      \cdots \}$, where $l'_{ij}$ is a link with thickness $t'_{ij}$.
      Then, normalize thickness of the links as $\sum_j t'_{ij} \!=\! 1$, 
      in each $r_i$.
\end{procedure}

%

\section*{Appendix B. \ Function of Paradigme\\
--- Spreading Activation Rules}

Each node $P_i$ of the semantic network {\it Paradigme} computes its
activity value $v_i(T\!+\!1)$ at time $T\!+\!1$ as follows:
\[
  v_i(T\!+\!1) = \phi\left({ R_i(T) + R'_i(T) \over 2} + e_i(T)\right),
\]
where $R_i(T)$ and $R'_i(T)$ are activity (at time $T$) collected from
the nodes referred in the r\'ef\'erant and r\'ef\'er\'e respectively; 
\ $e_i(T) \!\in\! [0,1]$ is activity given from outside (at time $T$); 
\ the output function $\phi$ limits the value to [0,1].

$R_i(T)$ is activity of the most {\it plausible} subr\'ef\'erant in 
$P_i$, defined as follows:
\[
  R_i(T) = S_{im}(T),
\]\[
  m = \mbox{argmax}_j \{ h_{ij} \!\cdot\! S_{ij}(T) \},
\]
where $h_{ij}$ is thickness of the $j$-th subr\'ef\'erant of $P_i$.
$S_{ij}(T)$ is the sum of weighted activity of the nodes referred in the
$j$-th subr\'ef\'erant of $P_i$, defined as follows:
\[
  S_{ij}(T) = \sum_k t_{ijk} \!\cdot\! a_{ijk}(T),
\]
where $t_{ijk}$ is thickness of the $k$-th link of $s_{ij}$, and 
$a_{ijk}(T)$ is activity (at time $T$) of the node referred by the 
$k$-th link of $s_{ij}$.

$R'_i(T)$ is weighted activity of the nodes referred in the 
r\'ef\'er\'e $r_i$ of $P_i$:
\[
  R'_i(T) = \sum_k t'_{ik} \!\cdot\! a'_{ik}(T),
\]
where $t'_{ik}$ is thickness of the $k$-th link of $r_i$, and 
$a'_{ik}$ is activity (at time $T$) of the node referred by the 
$k$-th link of $r_i$.

%

\end{document}